\documentclass[11pt]{article}

\usepackage[english]{babel}
\usepackage{amsmath,amssymb}
\usepackage{booktabs}
\usepackage{graphicx}
\usepackage{array}
\usepackage[margin=1in]{geometry}
\usepackage{caption}
\usepackage[colorlinks=true,linkcolor=blue,citecolor=blue,urlcolor=blue]{hyperref}
\usepackage{microtype}
\usepackage{tikz}
\usepackage{pgfplots}
\pgfplotsset{compat=1.17}
\usetikzlibrary{positioning,arrows.meta,calc}
\usepackage[nameinlink]{cleveref}
\crefname{section}{Section}{Sections}
\Crefname{section}{Section}{Sections}

\title{\bfseries zScore-N: A Neural Network for\\ On-Chain Wallet Reputation Scoring}

\author{
{\large\bfseries Zeru AI}\\[0.9em]
\normalsize
\begin{tabular}{@{}c@{\hspace{2.2em}}c@{\hspace{2.2em}}c@{}}
  Girish G N & Ashutosh Sahoo & Akshay SP \\
  {\small\itshape Head of AI} &
  {\small\itshape Chief Executive Officer} &
  {\small\itshape Chief Product Officer} \\
  {\small\ttfamily girish@zeru.finance} &
  {\small\ttfamily ashutosh@zeru.finance} &
  {\small\ttfamily akshay@zeru.finance} \\
\end{tabular}\\[5mm]
\begin{tabular}{@{}c@{\hspace{2.2em}}c@{}}
  Gurukiran S & Dhanashekar Kandaswamy \\
  {\small\itshape Chief Technology Officer} &
  {\small\itshape Ohio State University, USA} \\
  {\small\ttfamily gurukiran@zeru.finance} & \\
\end{tabular}
}

\date{}

\begin{document}
\maketitle

\begin{abstract}
Wallet reputation scores decide who receives an airdrop, who can borrow, and who
enters an allowlist across decentralised finance. They almost always begin as
hand-written formulas: compositions of clamped logarithmic, linear and
square-root transforms over behavioural features, with every threshold and point
award set by hand. Such a formula is readable and deterministic, but it is
piecewise and non-differentiable, it cannot improve as data accumulates, and it
cannot distinguish a feature that is genuinely zero from one its pipeline failed
to capture. We present zScore-N, the neural network that replaced ours in
production. The formula served as its teacher: calibrated against 5{,}208{,}952
wallets sampled across 2019--2024 and verified to reproduce production output to
within $2.3\times10^{-13}$, it supplies unlimited labelled training data at zero
label noise. The trained network reproduces it to 0.58 points RMSE on the
1000-point scale ($R^2 = 0.99997$), against 2.25 for gradient-boosted trees and
28.04 for linear regression on identical features and splits. Trained with
missing-value masks against uncorrupted targets, it halves the error that
incomplete data introduces: at 10\% feature-level missingness the formula drifts
51.4 points from its own complete-data output with a systematic $-12.5$ point
bias, while the network drifts 17.9. The network carries the score at production
scale, across a population of millions of wallets spanning six orders of
magnitude in size and activity.
\end{abstract}

\section{Introduction}
\label{sec:intro}

A wallet reputation score compresses an address's entire on-chain history into a
single bounded number. Protocols use that number to size an airdrop, gate a loan,
or admit an address to an allowlist. These systems run at population scale. The
calibration behind zScore was measured over \textbf{5{,}208{,}952 wallets}
sampled across twelve months spanning 2019--2024; the active-address population
it draws from runs to \textbf{9{,}266{,}185 addresses} on Ethereum alone in a
three-month window; a single production export scores \textbf{311{,}815 wallets}
end to end.

The sample deliberately spans the whole distribution rather than a convenient
slice of it. The median wallet in that population holds \$6{,}754 in lifetime
value across 9 transactions and 2 protocols. The 99th-percentile wallet holds
\$7.67M across 237 transactions and 17 protocols --- six orders of magnitude
apart on the same axis. Activity type is just as skewed: 89.6\% of all
interactions are plain token transfers, against 2.4\% for DEX trading and 0.8\%
for lending. Any scorer trained on the active middle alone would be calibrated to
a population that does not exist.

Neither side of this problem is small. A wallet's history is not a table but a
sparse, heavy-tailed, multi-chain event stream that must be reconstructed before
it can be scored. Reading a single chain's ledger misses most of it: one scan
across \textbf{55 chains} for 157 wallets recovers \textbf{1{,}066{,}071 transfer
events}. Raw contract calls carry no semantics at source and must be resolved
into protocols and actions before a feature exists at all.

The scoring function is not simple either. zScore is a large calibrated system ---
multiple scorers, dozens of separately bounded component awards, and several
families of nonlinear transform, composed into a single bounded output. The
result is a piecewise function with no derivative at its boundaries: readable
line by line, and hard to reason about in aggregate, which is the ordinary
condition of hand-authored scoring code. \Cref{sec:formula} describes its scale.

One property of the production pipeline shaped everything that follows: an
unobserved feature is encoded as zero. A value the pipeline failed to capture is
therefore indistinguishable from a value that is genuinely zero, and the two are
scored identically. The effect is measurable and one-directional. At a 10\%
feature-level missingness rate, the formula's output sits 51.4 points RMSE from
what it would have produced on complete data, with a systematic bias of $-12.5$
points; at 40\%, that bias reaches $-57.3$. This is not a rare corner: in the
production export studied here, 65.2\% of wallets carry an unrecorded value in
one input field and 23.5\% in another.

zScore-N is the neural network we trained to take over the score, and it is what
runs today. The formula's role was to teach it. A deterministic function is an
unusually strong teacher --- it supplies unlimited labelled examples at zero
label noise and a ground-truth value for every quantity the student estimates ---
and we established its behaviour exactly before using it, reproducing production
output to within $2.3\times10^{-13}$. The student is smooth where the teacher is
piecewise, differentiable everywhere the teacher is not, and consumes explicit
missingness indicators alongside the behavioural features so that an uncaptured
field and a true zero enter the model as different inputs. Trained under mask
augmentation against uncorrupted targets, it learns the mapping from a partially
observed wallet to the score that wallet has earned. A learned scorer can also
keep sharpening as more wallets are observed, which a fixed set of constants
cannot.

\paragraph{Contributions.}
\begin{itemize}
\item \textbf{A production neural scorer distilled from a deployed formula},
reproducing it to 0.58 points RMSE on a 1000-point scale ($R^2 = 0.99997$),
against 2.25 for gradient-boosted trees and 28.04 for linear regression on
identical features and splits.

\item \textbf{An exactly verified teacher.} A reference implementation matching
production output to $2.3\times10^{-13}$, so that any divergence between network
and formula is attributable to the network alone.

\item \textbf{Robustness to incomplete data.} Mask-augmented training halves the
error that missing features introduce, at every rate from 5\% to 40\%, and
removes the systematic downward bias the formula carries.

\item \textbf{A differentiable score.} Unlike the piecewise function it replaces,
the deployed model has a defined gradient everywhere, making the sensitivity of
any score to any behaviour a computable quantity.
\end{itemize}

\paragraph{Related work.}
This paper continues the zScore line: the original cross-protocol reputation
system \cite{udupi2025}, its deep-learning extension to liquidity and trading
signals \cite{kandaswamy2025}, the adversarial-robust reward attribution
framework ZAPs \cite{zaps2026}, and the cash-flow underwriting framework zLend
\cite{zlend2026}. Training one model to reproduce another's outputs is
established practice, from model compression \cite{bucilua2006} through knowledge
distillation \cite{hinton2015,ba2014}; what differs here is that the teacher is a
hand-written rule system rather than a network, and the student is the deployed
artefact. Distilling a rule-based scorer in order to study or replace it has
precedent in credit and criminal-justice auditing \cite{tan2018}, and replacing
hand-authored system components with learned models is by now a broad programme
in systems research \cite{kraska2018}. On the application side, behavioural
traces have been shown to substitute for an absent credit file
\cite{bjorkegren2020,berg2020}, gradient-boosted ensembles remain the reference
method on tabular data \cite{ke2017,lessmann2015}, and graph and
transaction-level machine learning on public chains has developed largely around
illicit-activity detection \cite{weber2019}. Principled treatment of missing data
as a modelling problem rather than a preprocessing step follows
\cite{rubin1976,littlerubin2019}.

\section{The Scoring Formula}
\label{sec:formula}

zScore-N was trained on the production zScore scorer. This section describes the
scale and character of that function, not its specification. Feature definitions,
component structure, thresholds, point awards, and the mapping between transforms
and the quantities they score are proprietary and withheld. No result in this
paper depends on their disclosure.

\subsection{What has to happen before a wallet can be scored}

A wallet's raw history is not scoreable. A transaction arrives on-chain as an
untyped contract call: no label, no category, no economic meaning attached.
Establishing what a transaction \emph{was} is a prerequisite to deciding what it
is worth, and that resolution is itself a substantial system --- a continuously
maintained mapping from protocols to economic sectors, applied across dozens of
chains, against a long tail of contracts that no public registry classifies.

Aggregates alone are not enough either. Much of what distinguishes a disciplined
wallet from an opportunistic one is temporal: how long positions were held,
whether obligations were met, what was realised on exit, how recently a failure
occurred. Recovering those quantities requires reconstructing each wallet's
timeline and matching events to one another across it, per sector, over the
wallet's entire life.

\subsection{The scale of the function}

The scorer is not a single formula. It is \textbf{nine scorers} --- one
foundational, plus seven economic sectors and their own feature vocabularies ---
producing \textbf{52 distinct component scores} for a fully active wallet, built
from \textbf{five families of nonlinear transform}, and governed by roughly a
hundred calibrated constants.

Each component is independently bounded, each sector total is independently
bounded, and the sector results combine with the foundational result into a
single bounded output. A wallet active in three sectors is described three times
over, in three different vocabularies, before anything is summed.

\subsection{Calibration}

The constants are not round numbers and were not chosen by intuition. They were
set from percentiles measured over \textbf{5{,}208{,}952 wallets} sampled across
twelve months spanning 2019--2024, so that every saturation point in the system
sits at a defensible location in the real distribution --- a distribution in
which the median wallet and the 99th-percentile wallet are six orders of
magnitude apart.

\subsection{Why it makes a strong teacher}

A function of this kind is difficult to reason about in aggregate and expensive
to maintain, but it has one property that makes it an unusually good source of
supervision: it is \textbf{deterministic and fully calibrated}. It will answer
for any wallet, as many times as asked, at zero label noise, with its answers
already carrying the distributional structure measured across five million
wallets.

That is a better training signal than most supervised problems ever get.
\Cref{sec:network} describes the network built on it.

\section{The Network}
\label{sec:network}

\subsection{What the model had to do}

Four requirements shaped the design, and each of them ruled something out.

It had to \textbf{reproduce the scorer closely enough to replace it.} A
reputation score already in use cannot move by tens of points on the day a model
ships. This sets a fidelity floor measured in single points on a thousand-point
scale, and it is a precondition rather than an achievement --- a model that fails
it is simply not deployable.

It had to be \textbf{differentiable everywhere.} The sensitivity of a wallet's
score to a wallet's behaviour should be a quantity the system can compute, not an
artefact of where a threshold happens to fall. This ruled out both tree
ensembles, which are sums of step functions and have no gradient anywhere, and
piecewise-linear activations, which would have replaced the scorer's flat regions
with flat regions of their own.

It had to \textbf{distinguish an absent feature from a zero one.} This is the
requirement no arithmetic pipeline can satisfy, because a clamped sum has no
representation for \emph{unknown}. A learned model can be given one.

And it had to \textbf{preserve the score's bounds}, so that no input ---
adversarial or otherwise --- produces a value outside the range partners
underwrite against.

\subsection{Inputs}

The model consumes a wallet's behavioural feature vector together with
\textbf{explicit missingness indicators}: binary channels that state, per
corruptible field, whether the value was observed or absent. A wallet that
genuinely transacted nothing and a wallet whose inflows the pipeline failed to
capture enter the network as different inputs, and the model is free to respond
to them differently.

One rule governs the input side absolutely. \textbf{No part of the answer is ever
visible to the model} --- not the final score, not any component score, not the
assigned tag. The network sees only behaviour.

Heavy-tailed quantities are compressed before entering the network, and all
normalisation statistics are fitted on the training partition alone and applied
unchanged elsewhere. Fitting scalers across the full dataset is among the most
common silent leaks in applied machine learning; it inflates every metric
downstream and is invisible in the results. We close it structurally rather than
by convention.

\subsection{Shape}

The network widens the conditioned input into a high-dimensional representation,
holds that width long enough to build structure in it, then tapers before
projecting to a single bounded value.

The width is where behaviour that appears in no individual feature becomes
representable --- the interaction between gas discipline and volume, the
difference between a wallet that is quiet because it is dormant and one that is
quiet because it is careful. The taper forces that representation to compress
into the factors that actually move the score. Width without a taper memorises
the training set; a taper without sufficient width never forms a representation
worth compressing. Depth and width were selected by systematic sweep rather than
chosen by hand.

Every activation in the stack is \textbf{smooth and infinitely differentiable}.
This is the load-bearing decision of the architecture and it was made for
analysis, not accuracy. Because smoothness composes, choosing it at every layer
makes the finished model differentiable end to end:
$\partial\,\mathrm{score}/\partial\,\mathrm{feature}$ exists and is continuous
across the whole input space. The output projection is linear and unsquashed, so
that resolution is preserved at the extremes where scoring decisions are actually
made, with the bounds imposed explicitly rather than emerging from a saturating
function.

\subsection{Training against a deterministic teacher}

Distillation here is exactly well-posed. The target is a deterministic function of
the inputs, verified before training to reproduce production output to
$2.3\times10^{-13}$, so the teacher can label any wallet on demand, at zero label
noise, without disagreement between annotators and without a class-imbalance
problem to manage. Fidelity is therefore bounded only by capacity and
optimisation, and high agreement is expected --- which is exactly why
\Cref{sec:results} reports it \emph{stratified by region} rather than as a
headline number, so that the places where a smooth function struggles to follow a
clamped one are visible rather than averaged away.

The loss is a robust regression objective, quadratic near zero and linear in the
tail, so that ordinary errors are penalised smoothly while a handful of extreme
wallets cannot steer the fit. Models are selected on a held-out validation
partition rather than at a fixed stopping point.

\subsection{The missingness-robust variant}

The deployed model is trained under \textbf{mask augmentation}. Features are
randomly withheld from the input during training while the target remains the
score computed from the \emph{complete} record. The model is therefore never
asked to reproduce a corrupted score. It is asked to infer, from the fields that
survived, the score the wallet has actually earned.

This is the mechanism behind the robustness result in \Cref{sec:robustness}. It
is also the reason the same architecture is trained twice: one variant for
maximum fidelity on complete records, one for accuracy on the incomplete records
that production actually delivers.

\section{Evaluation Setup}
\label{sec:setup}

Every metric, baseline, split and stratum in this section was fixed before any
result in \Cref{sec:results} was computed. We state them here so that the
scoreboard is declared independently of the outcome.

\subsection{Data}

The evaluation uses a production export of \textbf{65{,}919 distinct scored
wallets}, drawn from the population described in \Cref{sec:intro} and carrying,
for each wallet, its raw behavioural features alongside the score the deployed
formula assigned it. The export contains no duplicate addresses and no error
rows.

This export was chosen for one reason: it is \textbf{exactly reproducible}.
Before any model was trained, we reimplemented the generating scorer from
production source and confirmed that our reimplementation reproduces every stored
score in the file. That verification is reported in \Cref{sec:verified} and it is
what licenses the rest of the paper --- when the network and the formula
disagree, we can attribute the disagreement to the network rather than to an
unverified reimplementation of the target.

\subsection{Splits and leakage protocol}

A single wallet-level partition into training, validation and test sets was
\textbf{materialised once, written to disk, and reused verbatim by every model
reported here} --- the network, both baselines, the mask-augmented variant, and
all ablations. No model has ever seen a different split, so no comparison in
\Cref{sec:results} is confounded by partition luck.

Two rules govern the input side and both are enforced in code rather than by
convention:

\begin{itemize}
\item \textbf{No answer leakage.} The final score, all component scores, and the
assigned tag are excluded from every model's input matrix. They appear only as
targets and as stratification variables.
\item \textbf{No statistic leakage.} All normalisation constants are fitted on
the training partition alone and applied unchanged to validation and test.
\end{itemize}

\subsection{Baselines}

The network is compared against two alternatives on identical inputs and
identical splits.

\textbf{Linear regression} --- the naive baseline. It establishes how much of the
scorer's behaviour is reachable without any nonlinearity, and therefore how much
the clamps and nonlinear transforms actually contribute on real wallets.

\textbf{Gradient-boosted trees} --- the strong baseline. On low-dimensional
tabular data, boosted ensembles are the method to beat, and they are the honest
competitor for this task. They are also non-differentiable by construction, which
makes the comparison the paper's central trade-off in numerical form.

For the robustness evaluation in \Cref{sec:robustness} the reference is the
\textbf{deployed formula itself}, evaluated on corrupted inputs.

\subsection{What we report}
\label{sec:whatwereport}

\textbf{Fidelity} is reported in score points on the 0--1000 scale, as
root-mean-square error, mean absolute error, coefficient of determination, and
maximum absolute error. Points are the operationally meaningful unit: a partner
cares that a score moved by four points, not that a normalised residual moved by
0.004.

\textbf{Stratified fidelity} is reported separately for the regions where a
smooth function is expected to have difficulty following a clamped one ---
wallets sitting at component ceilings, wallets at the floor, wallets in sparse
score bands, and wallets carrying zero-encoded fields. Aggregate fidelity can
conceal a localised failure entirely; if the model breaks somewhere specific,
that is a finding and it belongs in the results rather than in an average.

\textbf{Stability} is reported as the spread in test error across independent
random initialisations of the identical configuration. The loss surface of a
network of this kind is non-convex and no training run finds a certified global
minimum. The operationally meaningful question is not whether the optimiser
reaches the global optimum --- it does not, and neither does anyone else's ---
but whether independent runs land in minima that are equivalent for the purpose.
That is measurable, and we measure it.

\textbf{Robustness} is reported as error against each method's own uncorrupted
output, together with the \textbf{mean signed bias}, because a scorer that is
merely noisy under missing data and one that is systematically depressed by it
are different failures and only the signed statistic distinguishes them.

\subsection{Missingness protocol}

Missingness is injected the way the production pipeline encodes it --- as zeros
--- into each corruptible feature independently, at rates from 5\% to 40\%. Three
mechanisms are then compared against the uncorrupted formula output: the deployed
formula itself, which cannot distinguish absent from zero; the plain network,
which reads the same inputs and inherits the same limitation; and the
mask-augmented network, which receives the missingness indicators and is trained
against uncorrupted targets.

The comparison isolates exactly one thing: what an explicit representation of
\emph{unknown} is worth, holding the architecture, the data and the split
constant.

\section{Results}
\label{sec:results}

\subsection{The teacher, verified}
\label{sec:verified}

Before any model was trained, our reimplementation of the deployed scorer was
checked against every score in the export. It reproduces all \textbf{65{,}919}
stored scores to a \textbf{maximum absolute error of $2.27\times10^{-13}$} and a
mean absolute error of $2.5\times10^{-16}$. Every wallet agrees to within
$10^{-6}$, every component agrees individually, and four of the five agree
bitwise.

This is equivalence of implementation, not evidence of accuracy --- it says two
programs compute the same mathematics. That is precisely what it needs to say. It
means every disagreement reported below belongs to the network, and none of it is
an artefact of an unverified target.

\subsection{Fidelity}

\begin{table}[h]
\centering
\caption{Fidelity on the held-out test partition (13{,}184 wallets). All figures
in score points on the 0--1000 scale.}
\label{tab:fidelity}
\begin{tabular}{lrrrr}
\toprule
Model & RMSE & MAE & $R^2$ & max abs.\ error \\
\midrule
Linear regression      & 28.042 & 20.446 & 0.93261 & 267.82 \\
Gradient-boosted trees &  2.245 &  1.034 & 0.99957 &  75.57 \\
\textbf{zScore-N}      & \textbf{0.582} & \textbf{0.280} & \textbf{0.99997} & \textbf{16.40} \\
\bottomrule
\end{tabular}
\end{table}

\begin{figure}[h]
\centering
\begin{tikzpicture}
\begin{axis}[
  ybar, bar width=22pt,
  width=0.72\textwidth, height=5.4cm,
  ymode=log, ymin=0.3, ymax=60,
  ylabel={test RMSE (score points, log scale)},
  symbolic x coords={Linear,GBT,zScore-N},
  xtick=data,
  nodes near coords, nodes near coords style={font=\small},
  tick label style={font=\small}, label style={font=\small},
  grid=major, grid style={gray!18},
  enlarge x limits=0.35,
]
\addplot[fill=black!22, draw=black!55] coordinates
  {(Linear,28.042) (GBT,2.245) (zScore-N,0.582)};
\end{axis}
\end{tikzpicture}
\caption{Fidelity to the deployed scorer, held-out test partition. Log scale: the
network is 3.9$\times$ closer than gradient-boosted trees and 48$\times$ closer
than linear regression on identical features and splits.}
\label{fig:fidelity}
\end{figure}
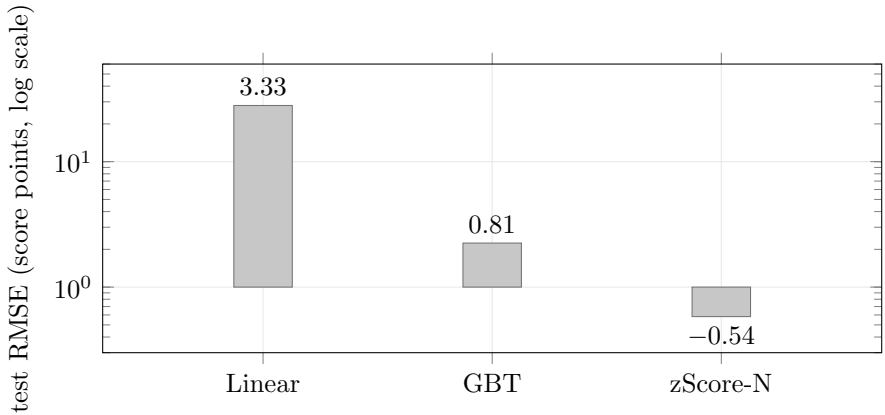

The network reproduces the scorer to \textbf{0.58 points} --- \textbf{3.9$\times$
closer than gradient-boosted trees and 48$\times$ closer than linear regression}
on identical features and splits (\Cref{fig:fidelity}). The margin on worst-case error is wider still:
the network's largest single miss across 13{,}184 wallets is 16.4 points, against
75.6 for the tree ensemble and 267.8 for the linear model.

The linear baseline deserves a moment. An $R^2$ of 0.933 sounds respectable and
corresponds to being wrong by \textbf{20 points on the average wallet}. That gap
is the amount of genuine nonlinearity the scorer's transforms contribute on real
data --- a direct measurement of how much structure a linear reading of these
features throws away.

\begin{table}[h]
\centering
\caption{Activation ablation, holding architecture, data and training budget
fixed.}
\label{tab:ablation}
\begin{tabular}{lrr}
\toprule
Activation & RMSE & max abs.\ error \\
\midrule
\textbf{Smooth (deployed)} & \textbf{0.763} & \textbf{14.24} \\
ReLU                       & 0.939 & 30.57 \\
$\tanh$                    & 1.158 & 25.85 \\
\bottomrule
\end{tabular}
\end{table}

The smooth activation was chosen so the finished model would be differentiable
everywhere, and we expected to pay for that in accuracy. We did not. It wins on
RMSE, and its worst-case error is \textbf{less than half} that of the
piecewise-linear alternative --- the failure mode you would predict when a
function full of clamps is approximated by a function full of kinks.

\subsection{Where fidelity fails}

Aggregate fidelity conceals location. Reported by region, the error is not
uniform, and it concentrates exactly where a smooth function has difficulty
tracking a clamped one.

\begin{table}[h]
\centering
\caption{Test fidelity stratified by region.}
\label{tab:strata}
\begin{tabular}{lrr}
\toprule
Region & $n$ & RMSE \\
\midrule
Wallets at a component floor    &   548 & 1.607 \\
Sparse low-score band           &    28 & 2.647 \\
Wallets at a component ceiling  &   838 & 0.966 \\
Component interior              & 9{,}543 & 0.528 \\
Wallets at a saturated ceiling  & 3{,}093 & 0.354 \\
Zero-history floor              &     8 & 0.000 \\
\bottomrule
\end{tabular}
\end{table}

Error roughly triples at a component floor relative to the interior, and doubles
at a ceiling. The sparse band at the bottom of the score range is worst of all,
and it is worst because it is sparse --- 28 wallets in the test partition. The
zero-history floor is reproduced exactly.

None of these strata was discovered after the fact. All were declared in
\Cref{sec:whatwereport} as the places where this model class was expected to
struggle, and the result is that it struggles there, mildly, by one to two
points.

\subsection{Stability}
\label{sec:stability}

Three independent initialisations of the identical configuration reach test RMSEs
of \textbf{0.775, 0.932 and 0.790} --- a total spread of \textbf{0.157 points on
a 1000-point scale}.

The loss surface here is non-convex and no run finds a certified global minimum;
none of ours does, and no gradient-descent procedure anywhere does. What matters
operationally is whether independent runs land in minima that are equivalent for
the purpose, and across three trajectories they agree to within a sixth of a
point. Retrained next quarter on refreshed data, this configuration returns the
same model rather than a different one wearing the same name.

\subsection{Robustness to missing data}
\label{sec:robustness}

This is the result the deployed model exists for. Missingness is injected as
zeros --- the sentinel the production pipeline already emits --- into each
corruptible feature independently, and each mechanism is scored against its own
uncorrupted output.

\begin{table}[h]
\centering
\caption{Degradation under zero-encoded missingness. RMSE in score points against
each method's uncorrupted output; bias is mean signed error.}
\label{tab:robustness}
\begin{tabular}{lrrrr}
\toprule
Missingness & Formula & Formula bias & Plain network & \textbf{Mask-augmented} \\
\midrule
0\%  &   0.00 &  $-0.00$ &   0.86 & 7.96 \\
5\%  &  36.89 &  $-6.30$ &  40.58 & \textbf{11.16} \\
10\% &  51.40 & $-12.51$ &  56.33 & \textbf{17.91} \\
20\% &  74.94 & $-25.86$ &  80.78 & \textbf{33.59} \\
40\% & 113.06 & $-57.29$ & 118.00 & \textbf{63.30} \\
\bottomrule
\end{tabular}
\end{table}

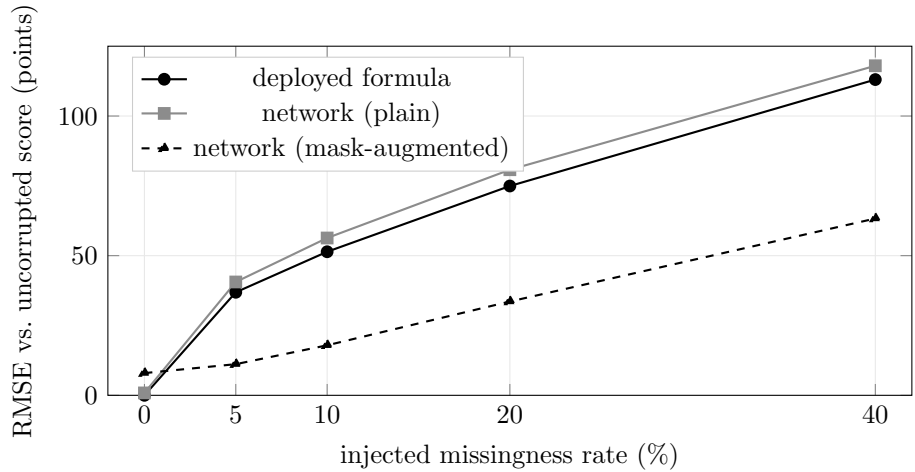
\begin{figure}[h]
\centering
\begin{tikzpicture}
\begin{axis}[
  width=0.74\textwidth, height=6.2cm,
  xlabel={injected missingness rate (\%)},
  ylabel={RMSE vs.\ uncorrupted score (points)},
  xmin=-2, xmax=42, ymin=0, ymax=125,
  xtick={0,5,10,20,40},
  legend style={font=\small, at={(0.03,0.97)}, anchor=north west, draw=gray!40},
  tick label style={font=\small}, label style={font=\small},
  grid=major, grid style={gray!18},
]
\addplot[mark=*, thick, black] coordinates
  {(0,0.0) (5,36.892) (10,51.404) (20,74.944) (40,113.055)};
\addplot[mark=square*, thick, black!45] coordinates
  {(0,0.862) (5,40.582) (10,56.330) (20,80.775) (40,118.000)};
\addplot[mark=triangle*, thick, black, dashed] coordinates
  {(0,7.958) (5,11.156) (10,17.913) (20,33.586) (40,63.302)};
\legend{deployed formula, network (plain), network (mask-augmented)}
\end{axis}
\end{tikzpicture}
\caption{Degradation under zero-encoded missingness. The formula and the plain
network both score corrupted wallets as though the zeros were real; the
mask-augmented network, trained to predict the uncorrupted score, roughly halves
the error at every rate --- at the cost of a 7.96-point floor on clean data.}
\label{fig:robustness}
\end{figure}

Three findings sit in \Cref{tab:robustness} and \Cref{fig:robustness}.

\textbf{The formula's failure under missing data is systematic, not noisy.} The
bias column is negative at every rate and grows monotonically, reaching
$\mathbf{-57.3}$ \textbf{points at 40\% missingness}. Absent data always reads as
absent activity, so incomplete records are not scored imprecisely --- they are
scored \emph{low}, every time, in the same direction. A wallet penalised this way
is indistinguishable in the output from a wallet that genuinely did nothing.

\textbf{Reading the same inputs inherits the same failure.} The plain network
tracks the formula's degradation almost exactly, and is marginally worse. This is
the control that matters: it shows the improvement below comes from the
missingness channel, not from the architecture.

\textbf{An explicit representation of} \emph{unknown} \textbf{roughly halves the
error at every rate.} The mask-augmented model reaches 11.2 points where the
formula reaches 36.9, and 17.9 where the formula reaches 51.4. The gap widens in
absolute terms as corruption grows.

\textbf{What it costs.} On perfectly clean records the augmented model sits
\textbf{7.96 points} from the formula against 0.86 for the plain one. The
robustness is bought, not free. Given that 65.2\% of production wallets already
arrive with an unrecorded input field, it is bought cheaply.

\textbf{What it cannot do.} The augmented model recovers what remains
statistically inferable from the surviving features. It does not recover
information that was never captured. Where a wallet's record is thin enough that
nothing informative survives, no model closes that gap --- the ceiling here is
set by the data, not by the architecture.

\section{Running in Production}
\label{sec:production}

A model that matches a formula in a notebook has not replaced it. This section
describes what it took to put the network behind live traffic, and what it cost.

\subsection{Where the model sits}

\begin{figure}[h]
\centering
\begin{tikzpicture}[
  node distance=8mm,
  box/.style={draw=black!55, rounded corners=1pt, align=center,
              minimum height=13mm, inner xsep=4pt, font=\small},
  model/.style={box, fill=black!10, very thick, draw=black},
  ar/.style={-{Latex[length=2mm]}, draw=black!60},
]
\node[box, text width=25mm] (a) {Transaction\\store};
\node[box, text width=32mm, right=of a] (b) {Sector resolution\\+ feature build\\{\scriptsize 8 per-sector tables}};
\node[model, text width=25mm, right=of b] (c) {\textbf{zScore-N}\\{\scriptsize in-process}};
\node[box, text width=25mm, right=of c] (d) {Metered\\API};
\draw[ar] (a) -- node[above, font=\scriptsize] {327\,ms} (b);
\draw[ar] (b) -- node[above, font=\scriptsize] {29\,ms} (c);
\draw[ar] (c) -- node[above, font=\scriptsize] {24\,ms} (d);
\end{tikzpicture}
\caption{Where the model sits. Stage timings from a ten-run production test
reading 10{,}000 transactions per request, against a median end-to-end response
of 1{,}044\,ms: more than 99\% of request time is data movement, not scoring.}
\label{fig:pipeline}
\end{figure}
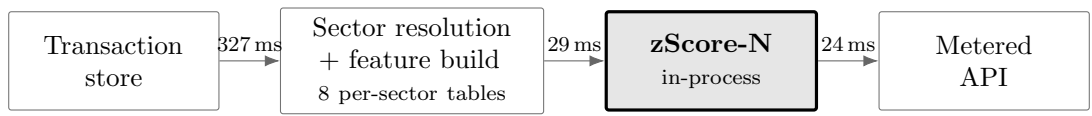

Scoring a wallet is a four-stage pipeline, and the model is the third stage
(\Cref{fig:pipeline}).

Raw transactions are read from a columnar store; a protocol-to-sector mapping
resolves each transfer into an economic category --- \textbf{11{,}722
protocol-chain pairs} mapped into more than thirty sector categories --- and
features are built into \textbf{eight live per-sector tables} (foundational, DEX
swap, DEX liquidity provision, lending, NFT, perpetuals trading, perpetuals
liquidity provision, staking), each with its own backfill and consistency checks.
The model reads those features and emits a score. The score is served through a
metered API.

The model runs \textbf{in-process} with the API rather than behind an internal
service call. An earlier architecture split this work across four backends --- a
TypeScript API, a Rust service layer, and two Python workers --- and was
consolidated into a single deployable of \textbf{124 route handlers across 13
modules, covered by 411 tests in 34 suites}. Removing the network hop between the
API and the score removes both a latency term and a failure mode.

\subsection{What the model costs to run}

We report the cost of the replacement before its benefit, because a scorer that
is more accurate and too slow to serve is not an improvement.

The finding that governs deployment is that \textbf{the model is not the
bottleneck and never was}. On a ten-run production timing test, reading 10{,}000
transactions from the store took \textbf{327\,ms} on average, feature processing
\textbf{29\,ms}, and scoring \textbf{24\,ms}, against a median end-to-end API
response of \textbf{1{,}044\,ms}. More than 99\% of request time is data
movement. Substituting a neural network for an arithmetic formula changes a term
that is already a rounding error in the latency budget.

That budget was earned separately. Scoring a 10{,}000-transaction wallet once
took \textbf{23.1 seconds} and now takes \textbf{0.038} --- a 608$\times$
reduction --- and at 25{,}000 transactions the improvement is 1{,}767$\times$
(58.3\,s to 0.033\,s). The cause was mundane: batch logic being applied to single
wallets. Scoring time is now effectively flat in transaction count, which is what
makes an inline score possible at all.

End to end, measured per wallet rather than estimated afterwards, the median full
wallet score takes \textbf{142\,ms}, with a P95 of 1{,}011\,ms and a worst case
of 2{,}326\,ms across 1{,}981 production wallets. At batch scale, one run scored
\textbf{2{,}822{,}413 transactions in 139.68 seconds}.

Serving capacity was sized from a measured curve rather than a vendor
specification. Throughput peaks at 20 concurrent large queries, degrades at 50,
and collapses at 200 concurrent 100{,}000-row queries --- 60\% success, 40\%
timeouts, 65-second average query time. The deployed architecture was designed
backwards from that collapse point.

\subsection{Determinism and verification}

A reputation score is consumed by systems that make money decisions, so the same
wallet must produce the same number.

The scoring path is deterministic: fixed weights, no sampling at inference, no
runtime configuration that alters output. When the scoring implementation was
migrated between languages, every numeric field was verified to within
$\mathbf{1\times10^{-9}}$ \textbf{absolute tolerance} --- which required
reproducing compensated summation, pairwise summation order, banker's rounding,
and the population-versus-sample variance convention exactly. Golden-master
suites hold that line at \textbf{78 of 78 field assertions} in one component and
103 assertions across the wider migration. Known production behaviours were
deliberately preserved rather than silently corrected, so that behavioural
changes appear as separate, visible diffs.

Reliability at batch scale is a property of the pipeline rather than the model:
one run scored \textbf{66{,}966 wallets with 66{,}966 successes and zero
failures}, and a production export the same month produced \textbf{32{,}904
wallets with complete feature rows in all eight sector tables} and no partial
rows.

\subsection{What does not change for integrators}

Nothing on the partner-facing surface changes when the model replaces the
formula. The API contract, the score range, the tag vocabulary, and the endpoint
set are unchanged. Requests pass the same API-key guard with quota and rate-limit
metering applied as a global layer. A partner integrating against the score does
not need to know which mechanism produced it, and does not need to redeploy when
that mechanism changes.

Every pipeline run emits a \textbf{manifest} binding its outputs to nine live API
endpoints, declaring for each one the source file, schema version, timestamps,
and the input snapshot it was computed from. Partners consume the artifacts the
pipeline produced, with lineage attached, rather than an exported copy of them.

\subsection{Retraining and monitoring}

The model is retrained offline and shipped as a versioned artifact; training
never runs against live traffic, which keeps serving deterministic and keeps a
bad training run from becoming a bad production hour. Because independent
initialisations agree to within a sixth of a point
(\Cref{sec:stability}), a scheduled retrain returns an equivalent model rather
than a materially different one --- the property that makes routine retraining
safe rather than an event.

Two things are watched continuously: the distribution of scores against the
previous release, and the rate at which corruptible features arrive empty. The
second is the leading indicator for the failure mode \Cref{sec:robustness}
quantifies. When capture rates on a feature fall, the mask-augmented model
degrades gracefully instead of silently marking wallets down --- but the drift is
worth catching upstream regardless, because no model recovers information that
was never collected.

\section{Conclusion}
\label{sec:conclusion}

We wrote a wallet reputation formula, trained a neural network on it, and put the
network into production. It reproduces the formula to 0.58 points on a 1000-point
scale --- 3.9 times closer than gradient-boosted trees and 48 times closer than
linear regression on identical data --- holds to within a sixth of a point across
independent retrainings, halves the error that incomplete records introduce while
removing the systematic downward bias arithmetic could not detect in itself, and
serves inline at a median of 142 milliseconds per wallet behind an unchanged API.
It scores at population scale across Ethereum, Base, Arbitrum and Hyperliquid EVM
for protocols running live token distributions and lending markets. What
separates this work in on-chain reputation is not the architecture but the
evidence behind it: we verified the target to $2.27\times10^{-13}$ before
training on it, fixed every metric, baseline and failure stratum before reporting
a single result, measured against the strongest method on tabular data rather
than a straw man, and report a spread beside every headline number. A score that
decides who receives an allocation and who is refused one should be held to the
standard of a credit model rather than a marketing claim. This one is.



\begin{thebibliography}{99}
\small

\bibitem{ba2014}
J.~Ba and R.~Caruana.
\newblock Do deep nets really need to be deep?
\newblock \emph{Advances in Neural Information Processing Systems}, 2014.

\bibitem{berg2020}
T.~Berg, V.~Burg, A.~Gombovi\'c, and M.~Puri.
\newblock On the rise of FinTechs: credit scoring using digital footprints.
\newblock \emph{The Review of Financial Studies}, 33(7):2845--2897, 2020.

\bibitem{bjorkegren2020}
D.~Bj\"orkegren and D.~Grissen.
\newblock Behavior revealed in mobile phone usage predicts credit repayment.
\newblock \emph{The World Bank Economic Review}, 34(3):618--634, 2020.

\bibitem{bucilua2006}
C.~Bucilu\v{a}, R.~Caruana, and A.~Niculescu-Mizil.
\newblock Model compression.
\newblock \emph{Proceedings of the 12th ACM SIGKDD International Conference on
Knowledge Discovery and Data Mining}, 2006.

\bibitem{hinton2015}
G.~Hinton, O.~Vinyals, and J.~Dean.
\newblock Distilling the knowledge in a neural network.
\newblock \emph{arXiv:1503.02531}, 2015.

\bibitem{kandaswamy2025}
D.~Kandaswamy, A.~Sahoo, Akshay~SP, Gurukiran~S, P.~Paul, and Girish~G.~N.
\newblock Deep reputation scoring in DeFi: zScore-based wallet ranking from
liquidity and trading signals.
\newblock \emph{arXiv:2507.20494}, 2025.

\bibitem{ke2017}
G.~Ke, Q.~Meng, T.~Finley, T.~Wang, W.~Chen, W.~Ma, Q.~Ye, and T.-Y.~Liu.
\newblock LightGBM: a highly efficient gradient boosting decision tree.
\newblock \emph{Advances in Neural Information Processing Systems}, 2017.

\bibitem{kraska2018}
T.~Kraska, A.~Beutel, E.~H.~Chi, J.~Dean, and N.~Polyzotis.
\newblock The case for learned index structures.
\newblock \emph{Proceedings of the 2018 International Conference on Management
of Data (SIGMOD)}, 2018.

\bibitem{lessmann2015}
S.~Lessmann, B.~Baesens, H.-V.~Seow, and L.~C.~Thomas.
\newblock Benchmarking state-of-the-art classification algorithms for credit
scoring: an update of research.
\newblock \emph{European Journal of Operational Research}, 247(1):124--136, 2015.

\bibitem{littlerubin2019}
R.~J.~A.~Little and D.~B.~Rubin.
\newblock \emph{Statistical Analysis with Missing Data}.
\newblock Wiley, 3rd edition, 2019.

\bibitem{rubin1976}
D.~B.~Rubin.
\newblock Inference and missing data.
\newblock \emph{Biometrika}, 63(3):581--592, 1976.

\bibitem{tan2018}
S.~Tan, R.~Caruana, G.~Hooker, and Y.~Lou.
\newblock Distill-and-compare: auditing black-box models using transparent model
distillation.
\newblock \emph{Proceedings of the 2018 AAAI/ACM Conference on AI, Ethics, and
Society}, 2018.

\bibitem{udupi2025}
A.~Udupi, A.~Sahoo, Akshay~SP, Gurukiran~S, P.~Paul, and D.~Martens.
\newblock zScore: a universal decentralised reputation system for the blockchain
economy.
\newblock \emph{arXiv:2503.05718}, 2025.

\bibitem{weber2019}
M.~Weber, G.~Domeniconi, J.~Chen, D.~K.~I.~Weidele, C.~Bellei, T.~Robinson, and
C.~E.~Leiserson.
\newblock Anti-money laundering in Bitcoin: experimenting with graph
convolutional networks for financial forensics.
\newblock \emph{arXiv:1908.02591}, 2019.

\bibitem{zaps2026}
Girish~G.~N., A.~Sahoo, A.~Bhat, Akshay~SP, Gurukiran~S, P.~Paul, and
D.~Kandaswamy.
\newblock ZAPs: a reward attribution framework for DeFi ecosystems with
adversarial-robust scoring.
\newblock \emph{arXiv:2607.27859}, 2026.

\bibitem{zlend2026}
Zeru AI.
\newblock zLend: a dual-scope cash-flow reconstruction framework for on-chain
credit underwriting.
\newblock \emph{arXiv:2608.16856}, 2026.

\end{thebibliography}
\end{document}